\documentclass[runningheads]{llncs}
\usepackage{booktabs}
\usepackage{stfloats}
\usepackage{amsfonts}
\usepackage[fleqn]{amsmath}
\usepackage{subcaption}
\usepackage[T1]{fontenc}
\usepackage{graphicx}
\usepackage{authblk}
\usepackage{etoolbox}
\makeatletter
\patchcmd{\thebibliography}
  {\clubpenalty4000\@clubpenalty\clubpenalty}
  {\clubpenalty4000\@clubpenalty\clubpenalty\setlength{\itemindent}{-2em}\setlength{\leftmargin}{2em}}{}{}
\makeatother

\begin{document}
\title{Attn-JGNN: Attention Enhanced Join-Graph Neural Networks}
%
%
\author{Jixin Zhang\orcidID{0009-0006-9758-7793}}
\authorrunning{J. Zhang et al.}
%
\institute{Key Labroatory of Symbolic Computation and Knowledge Engineering of Ministry of Education, Jilin University, Changchun, 130012, China\\
	\email{laiy@jlu.edu.cn}}
\maketitle              
\begin{abstract}
We propose an Attention Enhanced Join-Graph Neural Netw-orks(Attn-JGNN) model for solving \#SAT problems, which significantly improves the solving accuracy. Inspired by the Iterative Join Graph Propagation (IJGP) algorithm, Attn-JGNN uses tree decomposition to encode the CNF formula into a join-graph, then performs iterative message passing on the join-graph, and finally approximates the model number by learning partition functions. In order to further improve the accuracy of the solution, we apply the attention mechanism in and between clusters of the join-graphs, which makes Attn-JGNN pay more attention to the key variables and clusters in probabilistic inference, and reduces the redundant calculation. Finally, our experiments show that our Attn-JGNN model achieves better results than other neural network methods.
\end{abstract}

\section{Introduction}
Given a propositional formula, the model counting problem (\#SAT) aims to compute the number of satisfying assignments. As a fundamental problem in computer science, model counting has a wide range of practical applications, including probabilistic inference~\cite{DBLP:journals/ai/Roth96,DBLP:journals/ai/ChaviraD08}, probabilistic databases~\cite{DBLP:journals/ftdb/BroeckS17}, probabilistic programming~\cite{DBLP:journals/tplp/FierensBRSGTJR15}, neural network verification~\cite{DBLP:conf/ccs/BalutaSSMS19}, network reliability~\cite{DBLP:conf/ccs/BalutaSSMS19}. However, most of these model counting problems are \#P-hard~\cite{DBLP:journals/siamcomp/Valiant79}, posing significant challenges to computation. Although the scalability of exact model counters has been substantially improved, the inherent difficulty of this problem remains unchanged. Consequently, researchers have turned to exploring approximate methods to address model counting problems in real-world scenarios. The state-of-the-art approximate counting methods, such as ApproxMC~\cite{DBLP:conf/cp/ChakrabortyMV13}, satss~\cite{DBLP:journals/ai/GogateD11}, STS~\cite{DBLP:conf/uai/ErmonGS12}, PartialKC~\cite{DBLP:conf/aaai/LaiMY23}, have improved computational efficiency, yet they usually require invoking external SAT solvers in practical applications. Since the SAT problem itself is NP-hard, how to further enhance computational efficiency and scalability remains a major challenge in this field.

With neural networks demonstrating excellent learning abilities, various machine learning especially deep learning methods have been 
proposed for proposition model count~\cite{DBLP:conf/aaai/VaezipoorLWMGSB21,DBLP:conf/ijcnn/OzolinsFDGZK22,DBLP:journals/tkde/WuLL24}, including 
independent neural solver, directly predict the satisfaction of a given task distribution in implementing occuring
~\cite{DBLP:conf/iclr/AmizadehMW19,DBLP:journals/corr/abs-1903-01969}. Another research focus is to construct a general neural network 
framework by learning the approximate values of the partition function in statistical physics as an approximate \#SAT solver. This 
general network framework usually relies on propagation algorithms such as belief propagation algorithms. 
When the propagation algorithm converges, it corresponds to the critical point of Bethe free energy. The iterative process of the 
propagation algorithm is the process of finding the extreme point of bethe free energy~\cite{DBLP:journals/siamdm/ChandrasekaranCGSS11}. 
Our work is based on this framework. 

A recent work, NSNet~\cite{DBLP:conf/nips/LiS22}, a general graph neural network framework, describes the satisbility problem as a 
probabilistic reasoning problem on the graph, relying only on simple belief propagation (BP) as the message update rule in the latent 
space, and estimates the partition function to complete the approximate prediction. Encouraging results were shown on the \#SAT question.
However, although the BP algorithm is accurate in the tree structure, it inevitably generates repetitive messages when facing complex 
loop structures, resulting in NSNet being able to handle only specific graph structures and the solution accuracy being limited by the BP algorithm. 

Another kind of approximate model counter BPGAT~\cite{DBLP:conf/esann/Saveri22} by extending the BPNN architecture~\cite{DBLP:conf/nips/KuckCTLSSE20}, 
by introducing mechanism of attention, give important variables or higher weights of clause, Thereby improving the accuracy of understanding. 
However, due to the huge overhead brought by the global attention mechanism, it has not shown a very good effect on large-scale tasks, 
which is also limited by the graph structure. 

To solve the above problems, this paper proposes to use the Iterative join-graph Propagation (IJGP)~\cite{DBLP:conf/uai/DechterKM02}
algorithm combined with the attention mechanism~\cite{DBLP:journals/tkde/BrauwersF23,DBLP:conf/naacl/YangYDHSH16} to solve the \#SAT problem, 
which is called Attention Enhanced Join-Graph Propagation (Attn-JGNN). 
The IJGP algorithm is an approximate reasoning algorithm for probabilistic graphical models (such as Bayesian networks and Markov networks), 
aiming to effectively calculate the marginal probability or conditional probability of variables. The key idea is to approximate the precise 
solution by constructing a simplified join-graph and iteratively passing local messages. Compared with BP, IJGP can flexibly control the 
structure of the graph and the message-passing strategy by controlling the tree width of tree decomposition. 

We put the relevant variables and clause nodes into a clustering structure, connect different clusters through marked edges to form a join-graph, 
and apply the attention mechanism in each cluster of the join-graph to achieve a hierarchical effect. The Attn-JGNN model parameterizes the 
IJGP in the latent space through GNN and simulates its message update using the attention mechanism. IJGP avoids the repeated transmission of 
messages on the ring through edge marking, and its unique tree decomposition structure also enables us to better introduce the attention mechanism, 
thereby reducing the time complexity by an order of magnitude. Finally, similar to the previous framework, learn the partition function to approximately 
estimate the number of models. 

Specifically, in view of the hierarchical structure differences in message passing within and between clusters, we adopt a hierarchical structure 
where two attention layers are respectively responsible for message passing within and between clusters to improve the solution efficiency. We added 
a constraining awareness module in the loss function in the form of a regularization term, which prioritizes easily satisfied clauses and penalizes 
variable assignments that violate the constraints. Meanwhile, a dynamic attention mechanism is adopted. By dynamically increasing or decreasing the 
number of attention heads along with the time step, the training speed is improved and the resource consumption is reduced. 

In the ablation experiment, we proved that the above three improvements were effective. And IJGP is significantly superior to the BP algorithm. 
The experimental results on the BIRD and SATLIB benchmark datasets show that, with RMSE as the metric, compared with NSNet and BPGAT, the solution 
accuracy of Attn-JGNN has increased by 31\% and 45\% respectively.

This paper constructs a neural network framework Attn-JGNN. This framework applies the hierarchical attention mechanism to the join-graph of 
the IJGP algorithm and optimizes the framework through two methods: constraint awareness and dynamic trimming of the attention head. It breaks through 
the limitations of the graph structure imposed by traditional propagation algorithms and is more efficient when combined with attention.

\section{Preliminaries}
\subsection{Satisfiability Problems} 
In propositional logic, a Boolean formula is composed of Boolean variables and logical operators (e.g., negation (¬), conjunction ($\land$), and disjunction ($\lor$)). It is standard practice to represent Boolean formulas in Conjunctive Normal Form (CNF), which takes the form of a conjunction of clauses—where each clause is a disjunction of literals (a literal is either a variable or its negation).
Given a CNF formula, the SAT (Satisfiability Problem) asks whether there exists any variable assignment that satisfies the formula. In contrast, the goal of \#SAT (Propositional Model Counting Problem) is to count the total number of such satisfying assignments (also called "models").

\subsection{Iterative Join-Graph Propagation}
IJGP (Iterative Join-Graph Propagation) is an approximate inference algorithm designed primarily to compute marginal probabilities in probabilistic graphical models (e.g., Markov Random Fields (MRFs) and Bayesian Networks (BNs)). It constructs a join-graph and performs iterative message passing over this graph to efficiently approximate complex probability distributions.

For a given probabilistic graphical model, its joint probability distribution can be expressed as a product of factors:

\begin{equation}
P(X)=\left(\frac{1}{Z}\right)\prod\limits_{i=1}\limits^m\phi_i(C_i)
\end{equation}
where \(\phi_i(C_i)\) is a factor defined on a subset of variables \(C_i\subseteq X\), and\(\textbf{\textit{Z}}\) is the 
normalization constant (partition function).
A join-graph is a structure that decomposes the factor graph of the model into multiple clusters. Each cluster contains a set of variables and their associated factors. To ensure correctness of subsequent inference, the join-graph must satisfy two key properties:Coverage: Each factor \(\phi_i\) must be included 
in at least one cluster.Connectivity: For any two clusters that share a variable, there exists a path connecting them, and all clusters 
on the path contain the variable.

In this work, the join-graph is constructed using the external tree decomposition tool flow-cutter; the tree-width of the decomposition is controlled manually.
In the join-graph, a message is a function transmitted from a cluster \(C_i\) to another cluster \(C_j\), defined as:

\begin{equation}
m_{i\rightarrow j}(S_{ij})=\sum\limits_{C_i\backslash S_{ij}}\phi_i(C_i)\prod\limits_{k\in ne(i)\backslash j}m_{k\rightarrow i}(S_{ki})
\end{equation}
\(S_{ij}=C_i\bigcap C_j\) is the set of shared variables between clusters \(C_i\) and \(C_j\),  \(ne(i)\) is the set of neighboring 
cluster of \(C_i\) and \(\sum\limits_{C_i\backslash S_{ij}}\) denotes summation over variables in \(C_i\backslash S_{ij}\).
The core of IJGP is to approximate marginal probabilities via iterative message passing. The algorithm proceeds as follows:Initialize all messages \(m_{i\rightarrow j}\) to uniform distributions,For each cluster \(C_i\), compute the message \(m_{i\rightarrow j}\) for every neighboring cluster \(C_j\), then, Update messages until convergence or the maximum number of iterations is reached, finally, for each 
variable X, its marginal probability P(X) is proportional to the product of all messages in the clusters that contain X:

\begin{equation}
P(X)\propto\prod\limits_{X\in C_i}m_{i\rightarrow j}(S_{ij})
\end{equation}

\subsection{Graph Attention Networks}
The Graph Attention Network (GAT) is a graph neural network model that leverages attention mechanisms to dynamically aggregate information from neighboring nodes. Its key advantage is the ability to assign adaptive attention weights to different neighbors, capturing the relative importance of each node in the graph.
For a target node v and its neighbor \(u \in N(v)\)(where N(v)denotes the set of neighbors of v), the attention weight \(\alpha_{vu}\) (representing the importance of node u to node v)is defined as:

\begin{equation}
\alpha_{vu}=\frac{exp(LeakyReLU(a^T[Wh_v\mid\mid Wh_u]))}{\sum\limits_{k\in N(v)}exp(LeakyReLU(a^T[Wh_v\mid\mid Wh_k]))}
\end{equation}
where \(W\in \mathbb{R}^{d'\times d}\) a learnable weight matrix;\(a\in \mathbb{R}^{2d'}\) is a learnable attention vector; 
\(\mid\mid\) denotes vector concatenation; 
The updated representation \(h'_v\) of node v is obtained by weighted aggregation of information from neighboring nodes:

\begin{equation}
h'_v=\sigma(\sum\limits_{u\in N(v)}\alpha_{vu} Wh_u)
\end{equation}

\section{Methodology}
\begin{figure}[h]
\centering 
\includegraphics[width=1\textwidth]{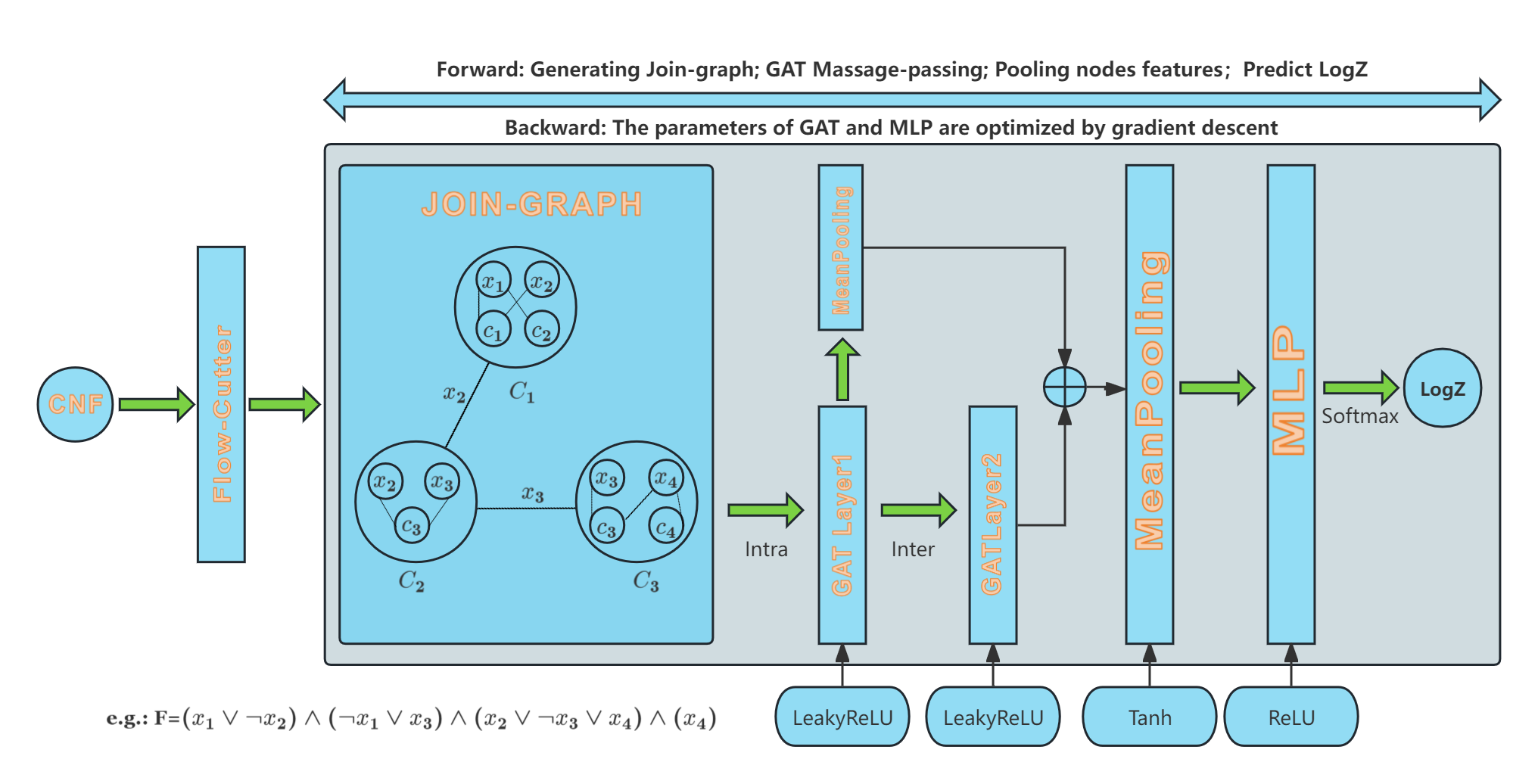}
\caption{For the \#SAT problem, our model uses two Graph Attention Network (GAT) layers for message passing and a Multi-Layer Perceptron (MLP) layer to estimate the partition function, serving as an approximate solver. A pooling layer compresses the processed variable and clause node features into a global representation, which is fed into the MLP layer.} \label{fig1}
\end{figure}

In this section, we first elaborate on the framework of our model (Attn-JGNN) and its operating principles, followed 
by an introduction to the integration of tree decomposition and attention mechanisms. As a neural-network-based 
implementation of the Iterative Join-Graph Propagation (IJGP) algorithm, this framework features a unique tree decomposition 
structure that facilitates better integration with the attention mechanism. By formulating \#SAT as a probabilistic 
inference task, we demonstrate how Attn-JGNN solves the problem (see Fig.~\ref{fig1}).
\subsection{Attn-JGNN Framework}
\begin{figure}[htbp]
    \centering
    \begin{subfigure}{0.3\textwidth}
        \centering
        \includegraphics[width=\linewidth]{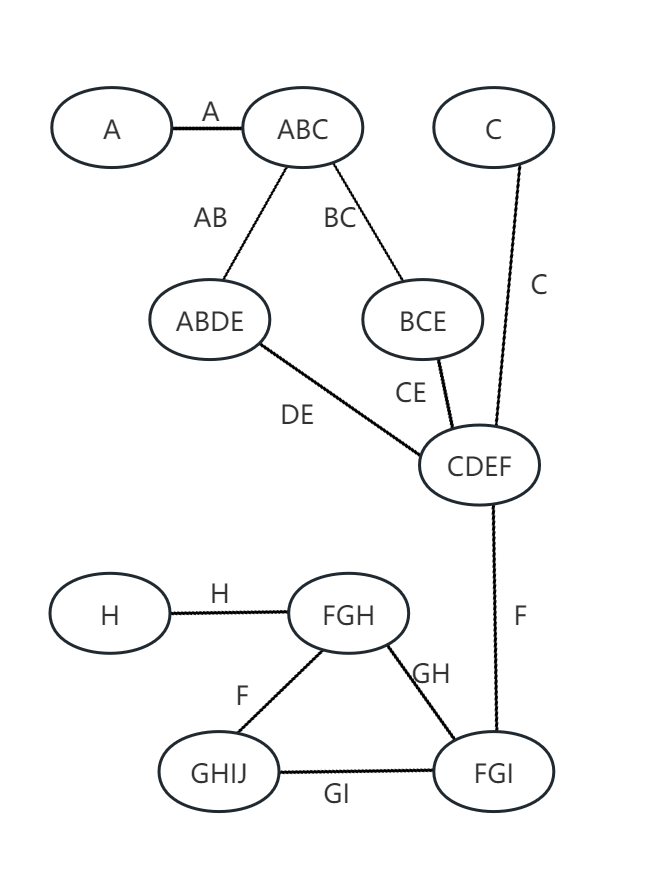} 
        \caption{}
        \label{fig:sub3}
    \end{subfigure}
    \begin{subfigure}{0.3\textwidth}
        \centering
        \includegraphics[width=\linewidth]{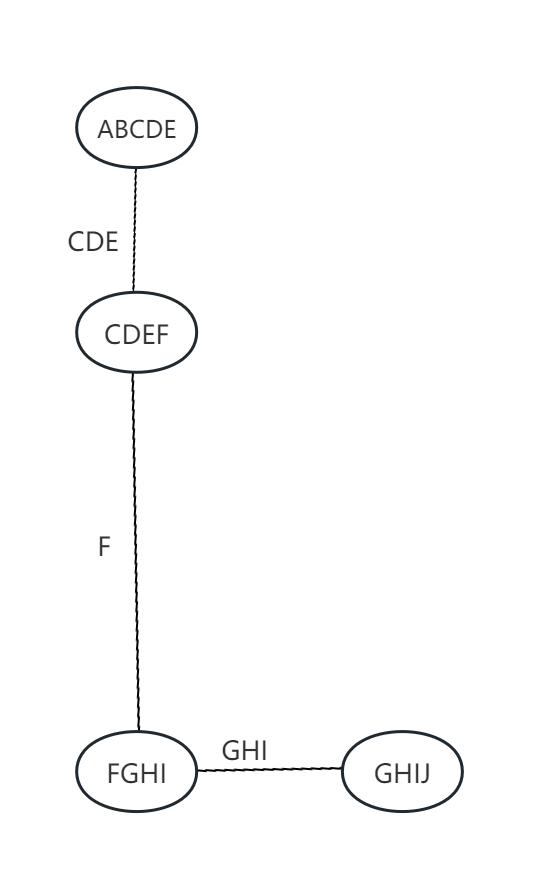} 
        \caption{}
        \label{fig:sub4}
    \end{subfigure}
    \caption{In the picture A,B,C... representing variables(Clause nodes are hidden, and clause nodes cannot appear on edges),the shared variables between the two clusters act as edge-lable. the same factor graph can be decomposed into different tree decomposition forms, figure (a) shows a low tree width but with poor accuracy, while figure (b) shows a high tree width, featuring high complexity but high accuracy}
    \label{fig2}
\end{figure}

For a given Conjunctive Normal Form (CNF) formula, we first encode it as a factor graph: an edge is established between 
a variable \(x_i\) and a clause \(C_j\) if \(x_i\) appears in \(C_j\), We then use an external tree decomposition tool 
to decompose this factor graph into a join-graph (consistent with the definition in Section 2.2), generating a set of 
clusters \{\(C_1\), \ \(C_2\), \ldots \(C_k\)\}. Each cluster contains variables and clauses that form a local substructure (see Fig.~\ref{fig2}).
At the input layer, we initialize two types of features:\(h_v\) and self-identifying node feature \(h_\phi\). The core 
architecture of the Attn-JGNN consists of two GAT layers(denoted GAT1 and GAT2), one MLP layer, and one pooling layer. 
\(GAT1\) and \(GAT2\) are cyclically invoked during message passing until convergence. \(GAT1\) is responsible for local 
variation-clause message passing, \(GAT2\) is responsible for cross-cluster message passing, and aggregates messages through 
splicing-pooling operation. Finally, \(b_i(C_i)\) and \(b_i(x_i)\) are estimated by the MLP layer to output the final 
number of models. The design of the architecture is in line with the iterative nature of the IJGP algorithm, that is, local 
first, then global, and the results within a cluster directly affect the propagation weight between clusters.

\subsection{Tree Decomposition and Attention}
Prior work has demonstrated the effectiveness of attention mechanisms for solving satisfiability problems~\cite{DBLP:conf/esann/Saveri22}. 
However, the high computational overhead of global attention limits scalability—for a CNF formula with n variables and m clauses,
global attention requires \(O((n+m)^2)\) computations to model interactions between all pairs of nodes. In Attn-JGNN, we address 
this issue by applying attention mechanisms per cluster (after tree decomposition of the factor graph). This reduces the computational 
complexity to \(O(kw^2)\), where k is the number of clusters and w is the maximum tree-width of the clusters. The advantage of 
this design becomes more pronounced as the problem scale increases.

We propose three tailored attention mechanisms to optimize Attn-JGNN, detailed below.In our work, we adopted three attention 
mechanisms to optimize the model, which are introduced in this section.In the Attention mechanism, Attention(Q,K,V) is the core 
computing module used to dynamically weight aggregated information based on the interaction of Query, Key, and Value. 
In Scaled Dot-Product Attention defined as:

\begin{equation}
Attention(Q,K,V)=softmax(\frac{QK^T}{\sqrt{d_k}})V
\end{equation}

\textbf{Hierarchical attention mechanism}
The hierarchical attention mechanism in Attn-JGNN aims to efficiently capture local and global dependencies in the graph via 
multi-granularity information aggregation. This design reduces computational overhead while enhancing the model’s ability to 
reason about complex constraints.

Local: The microscopic interaction between the attention-focused variable and the clause within the cluster (such as the polarity 
conflict of variables within the clause).The contribution weights of \(x_1\) and \(x_2\) to \(\phi_1\) are calculated in the 
cluster \(C_1=\{x_1,x_2,\phi_1=(x_1\bigvee \lnot x_2)\}\) so that high weights are assigned to variable assignments that are 
more likely to satisfy the clause; Variables and clauses inside cluster \(C\) calculate attention weights:

\begin{equation}
\alpha_{intra}=LekyReLU(\frac{(W_Qh_i)^T(W_Kh_j)}{\sqrt{d}}),\ \ \forall x_i,x_j\in C_k
\end{equation}
For variable node \(x_i\) and clause node \(\phi_j\) in cluster \(C\), the message passing formula is:
\begin{equation}
    m_{x_i\rightarrow \phi_j}^{(k)}=\alpha_{intra}\cdot \prod_{u\in \mathcal N(x_i)\backslash \phi_j}m_{u\rightarrow x_i}^{(k)}
\end{equation}

\begin{equation}
     m_{\phi_j\rightarrow x_i}^{(k)}=\alpha_{intra}\cdot\sum_{C_k \backslash \{x_i\}}\phi_j(C_k)\cdot\prod_{v\in \mathcal{N}(\phi_j)\backslash x_i}m_{v\rightarrow \phi_j}^{(k)}
\end{equation}
Update clause and variable feature:
\begin{equation}
h_j=\sum_{x_i\in C_k}\alpha_{intra}W_Vh_i
\end{equation}

Global: Inter-cluster attention transmits macro-constraints across clusters (such as consistency of assignment of distant variables) through shared variables.
If clusters \(C_1\) and \(C_2\) share the variable \(x_2\), then attention determines the influence of \(C_1\) and \(C_2\) on 
the assignment of \(x_2\). If \(C_1\) and \(C_2\) tend to conflict on \(x_2\), the attention weight automatically adjusts the 
message passing intensity.Calculate the attention weight of clusters \(C_1\)  to \(C_2\)  by passing cross-cluster messages 
through shared variables:

\begin{equation}
\alpha_{inter}=LekyReLU(\frac{(W_Qh_{C_1})^T(W_Kh_{C_2})}{\sqrt{d}})
\end{equation}
For adjacent clusters \(C_1\) and \(C_2\)(shared variable \(S_{12}=C_1\bigcap C_2\)), the inter cluster message is:
\begin{equation}
    m_{C_1\rightarrow C_2}(S_{12})=\alpha_{inter}\cdot\sum_{C_1\backslash S_{12}}(\phi_1(C_1)\cdot\prod_{k\in ne(C_1)\backslash C_2}m_{k\rightarrow C_1})
\end{equation}
Update shared variable characteristics:
\begin{equation}
h_x=h_x^{(C_1)}+\alpha_{inter}W_Vh_x^{(C_2)}
\end{equation}

\textbf{Dynamic attention mechanism}
The dynamic attention mechanism in Attn-JGNN model is realized by dynamically adjusting the number of attention heads to balance 
the performance of the model in different training stages and different complexity clauses. Start training with fewer attentional 
heads, quickly capture simple patterns (such as explicit constraints of short clauses), avoid overfitting, gradually increase the 
number of heads as the number of training steps increases to improve expressiveness, and deal with complex clauses (such as long chain dependencies) 

\begin{equation}
H(t)=min(H_{max},H_{init}+\lfloor\frac{t}{T}\rfloor)
\end{equation}
Assign a learnable weight to each attentional head \(\lambda_h\), dynamically adjusting its contribution:
\begin{equation}
\alpha_{dy}=\frac{1}{H(t)}\sum_{h=1}^H(t) \lambda_h Attention(Q,K,V)
\end{equation}
When \(\lambda_h\) is updated by gradient descent, the weight of important heads increases and the weight of redundant heads approaches 0.
This design allows Attn-JGNN to efficiently handle highly heterogeneous clause structures in \#SAT problems while maintaining low computational costs.

\textbf{Constraint-Aware Mechanism}
In Attn-JGNN, the central role of the Constraint-Aware Mechanism is to explicitly guide the model to preferentially satisfy 
clause constraints in the CNF formula, thus more efficiently approaching the correct model count. The realization method combines 
attention weight adjustment and loss function regularization.For each clause \(C_i\), define its satisfaction score \(s_i\):

\begin{equation}
s_i=sigmoid(\sum_{x_j\in \phi_i}(2b_j(x_j)-1)polarity(x_j,\phi_i))
\end{equation}
where, \(b_j(x_j)\) is  the current assignment probability of \(x_j\) ;  \(polarity(x_j,\phi_i)\) represents the polarity of \(x_j\) in the clause \(\phi_i\) .
\(s_i\in (0,1)\), where the closer to 1 means that the clause \(\phi_i\) is more likely to be satisfied.Add the following regularization terms to the loss function:
\begin{equation}
\mathcal L_{cons}=-\delta  \sum_{i=1}^mlns_i,
\end{equation}
Combining the RMSE and the constrained aware regularization term, the total loss function is:
\begin{equation}
\mathcal L_{total}=\mathcal L_{RMSE}+\mathcal L_{cons}
\end{equation}
The constraint awareness mechanism acts on the other mechanisms, implicitly adjusting the message passing process, using \(s_i\) 
weighted messages when propagating within and between clusters:
\begin{equation}
\alpha_{intra}=LekyReLU(\frac{(W_qh_i)^T(W_kh_j)+\gamma s_i}{\sqrt{d}})
\end{equation}

\subsection{\#SAT}
In join-graph, we need to modify the Bethe formula to fit the specific structure of the join-graph:

\begin{equation}
   F_{Bethe-Join}=\sum_\alpha [H(b_{C_\alpha})-\sum_{v\in C_\alpha}(d_v^\alpha-1)H(b_v)]
\end{equation}
\(H(b_{C_\alpha})\) is the joint distribution entropy of variables and clauses within cluster \(C_\alpha\), \(H(b_v)\) is the 
entropy of the local variable, are the \(GAT1\) and \(GAT2\) outputs respectively, which are used as the input of the MLP layer 
after the pooling operation. the goal of the MLP is to approximate \(F_{Bethe-Join}\) by  the hierarchical structure of the 
join-graph.Its inputs and specific implementation are as follows:
\begin{equation}
    h_{C_\alpha}=[H(b_{C_\alpha}), \sum_{v\in C_\alpha}(d_v^\alpha-1)H(b_v)], h_G=\frac{1}{\mid{C_\alpha}\mid}\sum_\alpha h_{C_\alpha}
\end{equation}

\begin{equation}
   H(b_{C_\alpha})=GAT1(\frac{1}{ \mid C_\alpha\mid}\sum_{j\in C_\alpha}h_j), H(b_v)=GAT2(h_x)
\end{equation}
The MLP fits the following mappings:

\begin{equation}
    \hat{F}_{Bethe-Join}=W_2\cdot ReLU(W_1h_G+b_1)+b_2
\end{equation}
\( W_1\in\mathbb{R}^{d\times 2}, b_1\in \mathbb{R^d}\) is the MLP hidden layer parameter\
and the \(W_2\in\mathbb{R}^{1\times d} , b_2\in \mathbb{R}\) is the output layer parameter. By supervised ground truth logZ 
(precomputed by the exact method), the loss function is designed as follows \(\mathcal L_{total}\), finally, make a prediction:
\begin{equation}
    logZ\approx -\hat{F}_{Bethe-Join}=-MLP(h_G)
\end{equation}

\section{Experimental Evaluation}
\subsection{Experiment Setup}
In all experiments, we set the feature dimension d=64 and the number of message-passing iterations T=5 for training. The model architecture consists of two Graph Attention Network (GAT) layers followed by a Multi-Layer Perceptron (MLP) layer (in a sequential connection). The initial number of attention heads is 4, and this number increases by 1 every 1000 training steps until reaching a maximum of 8. All experiments are conducted on a server equipped with a single NVIDIA A100 GPU and 8 CPU cores.

We first follow the experiment settings in recent work NSNet. Specifically, we run experiments using the 
same subset of BIRD benchmark~\cite{DBLP:conf/aaai/SoosM19} , which contains eight categories arising from DQMR networks, grid 
networks, bit-blasted versions of SMTLIB benchmarks, and ISCAS89 combinatorial circuits. Each category has 
20 to 150 CNF formulas, which we split into training/testing with a ratio of 70\%/30\%. Note that the BIRD 
benchmark is quite small and contains large-sized formulas with more than 10,000 variables and clauses, 
it challenges the generalization ability of our model. Besides evaluating in such a data-limited regime, 
we also conduct experiments on the SATLIB benchmark, an open-source dataset containing a broad range of CNF 
formulas collected from various distributions. To train our model effectively, we choose the distributions 
with at least 100 satisfiable instances, which include the following 5 categories: (1) uniform random 3-SAT 
on phase transition region (RND3SAT), (2) backbone-minimal random 3-SAT (BMS), (3) random 3-SAT with controlled 
backbone size (CBS), (4) "Flat" graph coloring (GCP), and (5) "Morphed" graph coloring (SW-GCP). The whole 
dataset has 46,200 SAT instances with the number of variables ranging from 100 to 600, and we split it into 
training/validation/testing sets with a ratio of 60\%/20\%/20\%. For both BIRD and SATLIB benchmarks, we ran 
the state-of-the-art exact \#SAT solver DSharp~\cite{DBLP:conf/ai/MuiseMBH12} with a time limit of 5,000 seconds to generate the 
ground truth labels. The instances where DSharp fails to finish within the time limit are discarded.

\subsection{Evaluation \& Baselines}
Following BPNN and NSNet, we use the (1) root mean square error (RMSE) between the estimated log countings and 
ground truth as our evaluation metrics. We compare Attn-JGNN , the neural baseline BPNN and NSNet, and two state-of-the-art 
approximate model counting solvers, ApproxMC3 and F2~\cite{DBLP:conf/sat/AchlioptasHT18a}. For ApproxMC3 and F2, we set a time 
limit of 5,000 seconds on each instance.

\subsection{Main Results}
\begin{figure}[htbp]
    \centering
    \begin{subfigure}{0.48\textwidth}
        \centering
        \includegraphics[width=\linewidth]{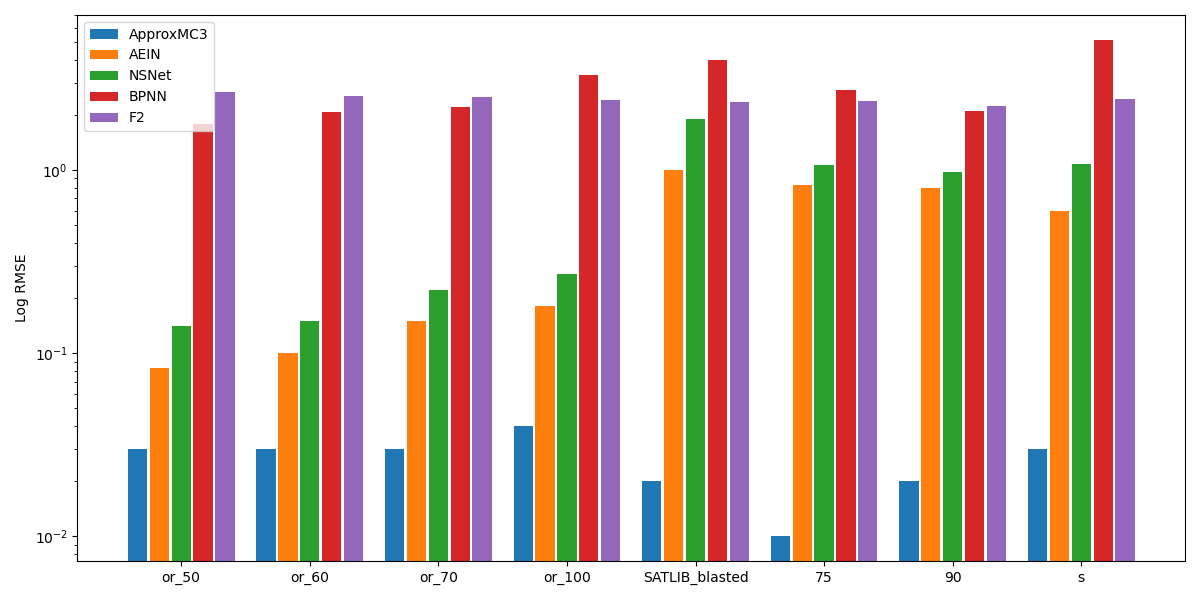} 
        \caption{}
        \label{fig:sub1}
    \end{subfigure}
    \hfill 
    \begin{subfigure}{0.48\textwidth}
        \centering
        \includegraphics[width=\linewidth]{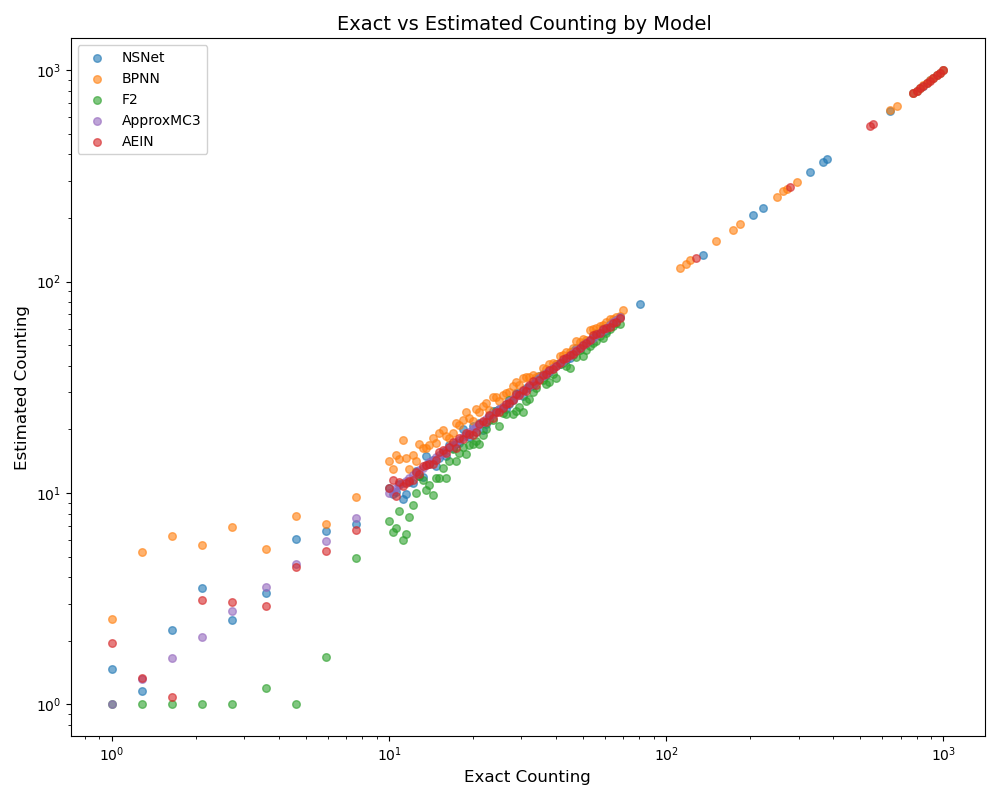} 
        \caption{}
        \label{fig:sub2}
    \end{subfigure}
    \caption{(a) is RMSE between estimated log countings and ground truth for each solver on the BIRD benchmark;(b) is Scatter plot comparing the estimated log countings against the ground truth for each solver on the BIRD benchmark}
    \label{fig:total}
\end{figure}

As shown in Figure~\ref{fig:sub1}, Attn-JGNN can estimate tighter counts than NSNet, BPNN, and F2 in all categories 
of the BIRD benchmark. Attn-JGNN estimates are almost three times more accurate than F2 and BPNN. However, Attn-JGNN cannot 
compete with ApproxMC3.

Figure~\ref{fig:sub2} shows the scatter plot. The estimated logarithmic count is compared to the ground truth 
for each solver on the BIRD benchmark. When the ground truth is less than \(e^{100}\), Attn-JGNN and ApproxMC3 can 
provide more accurate estimates than NSNet, F2 and BPNN in most cases. ApproxMC3 is unable to complete in 5000 
seconds when the ground truth count exceeds \(e^{100}\), Attn-JGNN can still give a close approximation when the ground 
truth count exceeds \(e^{1000}\). This demonstrates the effectiveness of Attn-JGNN in solving difficult and large cases.

The solution speed of Attn-JGNN without using the attention mechanism is same order of magnitude as that of NSNet , 
and its effect is still better than that of NSNet. This further indicates that the reasoning ability of the IJGP 
algorithm is superior to that of BP.

\begin{table}[htbp] 
  \centering  
  \caption{Comparison of RMSE between Attn-JGNN without attention mechanism and NSNet}  
  \begin{tabular}{ccclll}  
    \toprule
    Method& RND3SAT& BMS & CBS& GCP&SW-GCP\\  
    \midrule
    NSNet& 1.57& 2.45& 1.68& 2.14&1.37\\  
    Attn-JGNN-Att& \textbf{1.42}& \textbf{2.29}& \textbf{1.33}& \textbf{2.08}&\textbf{1.25}\\  
    \bottomrule
  \end{tabular}
  \label{tab1}  
\end{table}
Table~\ref{tab2} shows the detailed RMSE results for each solver on the SATLIB benchmark. The data for the BIRD benchmark 
is collected from many real-world model counting applications that may share many common logical structures to learn, whereas 
instances in the SATLIB benchmark are randomly generated, making it difficult for Attn-JGNN to exploit common features. Despite this, 
Attn-JGNN still outperforms NSNet and F2 in most categories.
\begin{table}[htbp] 
  \centering  
  \caption{RMSE between estimated log countings and ground truth for each solver on the SATLIB benchmark.}  
  \begin{tabular}{ccclll}  
    \toprule
    Method& RND3SAT& BMS & CBS& GCP&SW-GCP\\  
    \midrule
    F2& 2.13& 2.42& 2.37& 2.40&2.66\\  
    NSNet& 1.57& 2.45& 1.68& 2.14&1.37\\  
    Attn-JGNN& \textbf{1.15}& \textbf{1.66}& \textbf{1.20}& \textbf{1.96}&\textbf{0.96}\\  
    \bottomrule
  \end{tabular}
  \label{tab2}  
\end{table}
\begin{table}[htbp] 
  \centering  
  \caption{Ablation experiments of the Attn-JGNN model on three refinements.}  
  \begin{tabular}{ccclll}  
    \toprule
    Method& RMSE& Head utilization(\%)& Training time/convergence\\ 
    \midrule
    GAT& 1.33& 100& 185.155\\  
    GAT-H& 1.26& 100& 153.499\\  
    GAT-HC& 1.19& 100& 170.164\\
    \centering  
    GAT-HCD& 1.16& 62.5& 113.165\\  
    \bottomrule
  \end{tabular}
  \label{tab3}  
\end{table}
To prove that the three attention mechanisms of our Attn-JGNN model are effective, Table~\ref{tab3} shows its ablation experiments with 
RMSE, attention head utilization and training time as evaluation metrics. Experiments show that the three attention mechanisms all 
play a positive role in the model. Among them, the hierarchical attention mechanism and constraint perception mechanism greatly 
improve the accuracy of the model, and the dynamic attention mechanism reduces the redundant attention head over time, avoids more 
calculations, reduces the training time and improves the efficiency of the model.
\section{Related Works}
Since \#SAT was proven to be a \#P-complete problem, developing efficient solutions for \#SAT 
with limited computational resources has become a key research focus. Traditional model counting methods are categorized 
into two groups based on the required accuracy of results: exact counting and approximate counting. Recent advances have 
also introduced data-driven neural network approaches, which leverage learning capabilities to address \#SAT’s inherent
complexity.

Exact counting methods prioritize absolute correctness of results, making them suitable for scenarios with small variable 
scales or specialized formula structures. They can be further divided into search-based and dynamic programming (DP)-based 
approaches, depending on their core reasoning mechanisms.
Search-based methods typically extend the Davis-Putnam-Logemann-Loveland (DPLL) algorithm—an iterative search procedure for 
propositional satisfiability—to count satisfying assignments. While these methods guarantee exact results, their scalability 
is sometimes limited due to exponential time complexity in the worst case.
Well known tools in the search-based category includes c2d~\cite{DBLP:conf/ecai/Darwiche04}, SharpSAT~\cite{DBLP:conf/sat/Thurley06}, D4~\cite{DBLP:conf/ijcai/LagniezM17}, Ganak~\cite{DBLP:conf/ijcai/SharmaRSM19}, ExactMC~\cite{DBLP:conf/aaai/LaiMY21}, Panini~\cite{DBLP:conf/cav/LaiMY25}, etc. 
DP-based exact counters avoid brute-force search by decomposing the formula into subproblems and solving them recursively. Two 
representative methods are ADDMC~\cite{DBLP:conf/aaai/DudekPV20} and DPMC~\cite{DBLP:conf/cp/DudekPV20}.

Approximate counting methods trade off result accuracy for polynomial-time complexity, addressing the scalability gap of exact methods 
for large-scale CNF formulas. The most mainstream approaches in this category are hash-based approximate counters, which rely on 
randomization to estimate model counts without exhaustive enumeration.The core idea of hash-based methods is to partition the solution 
space (all variable assignments) into disjoint, uniformly sized "cells" using random hash functions.  The total number of models is 
then estimated by: (1) randomly selecting a cell;  (2) exactly counting the number of satisfying assignments within that cell;  
and (3) scaling the count by the total number of cells.
A pioneering and widely used solver in this area is ApproxMC~\cite{DBLP:conf/cp/ChakrabortyMV13} and its subsequent optimizations~\cite{DBLP:conf/ijcai/ChakrabortyMV16,DBLP:conf/aaai/SoosM19,DBLP:conf/cav/SoosGM20,DBLP:conf/cav/YangM23}. 
ApproxMC introduces random XOR constraints to partition the solution space—each XOR constraint defines a hash function that groups assignments
into cells. It provides provable approximation guarantees by controlling the number of XOR constraints and the number of sampled cells. 
However, ApproxMC’s performance heavily depends on the efficiency of its underlying SAT solver (used to count assignments in sampled cells) 
and requires careful engineering for state management and solver interaction.
While ApproxMC provide guaranteed approximation, there are also some efficient approximate model counters without guarantee, such as STS~\cite{DBLP:conf/uai/ErmonGS12}, satss~\cite{DBLP:journals/ai/GogateD11}, and PartialKC~\cite{DBLP:conf/aaai/LaiMY23}.

With the rise of deep learning, data-driven neural network approaches have emerged as a new paradigm for \#SAT, leveraging graph neural networks 
(GNNs) and message-passing architectures to learn patterns from formula structures. These methods do not rely on handcrafted heuristics, making 
them more adaptable to diverse formula distributions.
Early works focused on predicting satisfiability (SAT) rather than counting models. A foundational example is NeuroSAT~\cite{DBLP:conf/iclr/SelsamLBLMD19}, 
which uses a GNN to perform message passing on a variable-clause bipartite graph (nodes represent variables/clauses, edges represent membership). 
NeuroSAT learns to classify formulas as satisfiable or unsatisfiable by updating node features through iterative message exchange—demonst rating 
that neural networks can capture logical dependencies without explicit rule-based reasoning.
Recent works extend neural approaches to \#SAT by integrating message-passing algorithms (e.g., belief propagation) with neural networks, Proposed 
by Kuck et al., BPNN~\cite{DBLP:conf/nips/KuckCTLSSE20} combines belief propagation (BP) with a neural network architecture. It frames model counting 
as a probabilistic inference problem and uses BP to propagate beliefs (assignment probabilities) in the latent space. BPNN achieves up to 100x faster 
counting than state-of-the-art handcrafted solvers for certain formula classes, though it relies on BP’s limitations (e.g., inaccuracy on cyclic graphs).
Developed by Averi et al. , BPGAT~\cite{DBLP:conf/esann/Saveri22} extends BPNN by introducing an attention mechanism. It assigns higher weights to 
critical variables and clauses, enhancing the model’s ability to capture impactful logical constraints. BPGAT serves as an approximate model counter, 
improving accuracy over BPNN but suffering from high computational overhead due to global attention.

\section{Conclusions}
In this work, we propose a neural framework for solving the satisfiability problem. Our framework combines tree decomposition and GAT, includes IJGP in the latent space, and performs partition function estimation to solve \#SAT. Experimental evaluation on synthetic datasets and existing benchmarks shows that our approach significantly outperforms NSNet and other neural baselines and achieves competitive results compared to state-of-the-art solvers.
\begin{credits}
\subsubsection{\ackname} 
This work was supported in part by Jilin Provincial Natural Science Foundation [20240101378JC], Jilin Provincial Education Department Research Project [JJKH20241286KJ], and the National Natural Science Foundation of China [U22A2098, 62172185, and 61976050]

\end{credits}

\bibliographystyle{plain}
\bibliography{references}
\end{document}